\documentclass[journal]{IEEEtran}
\usepackage{graphicx}%
\usepackage{amsmath,amssymb,amsfonts}%
\usepackage{amsthm}%
\usepackage{mathrsfs}%
\usepackage{xcolor}%
\usepackage{textcomp}%
\usepackage{manyfoot}%
\usepackage{booktabs}%
\usepackage{algorithm}%
\usepackage{algorithmicx}%
\usepackage{algpseudocode}%
\usepackage{listings}%
\usepackage{stfloats}
\usepackage{verbatim} 
\usepackage{multirow}
\usepackage{hhline}
\usepackage{tabulary}
\usepackage{multicol}
\usepackage{tikz}
\usepackage{bbm, dsfont}
\usepackage{ulem}
\usepackage{epsfig}
\usepackage{bbding}
\usepackage{pifont}
\usepackage{booktabs}
\usepackage[colorlinks,linkcolor=blue]{hyperref}
\begin{document}
%\title{Graph-based Selective Filtering Network for Referring Expression Comprehension}
 \bstctlcite{IEEEexample:BSTcontrol}
     \title{Make Graph-based Referring Expression Comprehension Great Again through Expression-guided Dynamic Gating and Regression}
     \author{Jingcheng Ke,
       Dele Wang,
       Jun-Cheng Chen,~\IEEEmembership{Member,~IEEE,}
       I-Hong Jhuo,~\IEEEmembership{Member,~IEEE,}
       Chia-Wen Lin,~\IEEEmembership{Fellow,~IEEE,}
       and Yen-Yu Lin,~\IEEEmembership{Senior Member,~IEEE}
%  \thanks{Manuscript received June 30, 2024; revised August 06, 2024; accepted September 02, 2024. Date of publication month day, 2024; date of current version month day, 2024. This work was funded in part by National Science and Technology Council (NSTC) under grants 112-2634-F-002-005, 12-2221-E-A49-090-MY3, and in part by Qualcomm Technologies, Inc., through a Taiwan University Research Collaboration Project, under Grant NAT-487844-4.  (Corresponding author: Chia-Wen Lin)}     
%      \thanks{J. Ke, D. Wang, and C.-W. Lin are with the Department of Electrical Engineering and the Institute of Communications Engineering, National Tsing Hua University, Hsinchu 300044, Taiwan  (e-mail: freedom6927@gmail.com;alfred@m107.nthu.edu.tw; cwlin@ee.nthu.edu.tw).}% <-this % stops a space
%      \thanks{J.-C. Chen is with the Research Center for Information Technology Innovation, Academia Sinica, Taipei 115201, Taiwan (e-mail: pullpull@citi.sinica.edu.tw).}% <-this % stops a space
%      \thanks{I.-H. Jhuo is with Microsoft, Seattle, WA, USA (e-mail: ihjhuo@gmail.com).}
%      \thanks{Y.-Y. Lin is with the Department of Computer Science, National Yang Ming Chiao Tung University, Hsinchu 300093, Taiwan (e-mail: lin@cs.nycu.edu.tw).}% <-this % stops a space
       }% <-this % stops a space

%\markboth{IEEE Transactions on multimedia}{Ke \MakeLowercase{\textit{et al.}}: dynamic gate constraint (DGC)}

% ====================================================================
\maketitle

% === ABSTRACT ====================================================================
% =================================================================================
\begin{abstract}
One common belief is that with complex models and pre-training on large-scale datasets, transformer-based methods for referring expression comprehension (REC) perform much better than existing graph-based methods.
We observe that since most graph-based methods adopt an off-the-shelf detector to locate candidate objects (\textit{i.e.}, regions detected by the object detector), 
they face two challenges that result in subpar performance: (1) the presence of significant noise caused by numerous irrelevant objects during reasoning, and (2) inaccurate localization outcomes attributed to the provided detector.
To address these issues, we introduce a plug-and-adapt module guided by sub-expressions, called dynamic gate constraint (DGC), which can adaptively disable irrelevant proposals and their connections in graphs during reasoning. We further introduce an expression-guided regression strategy (EGR) to refine location prediction.
Extensive experimental results on the RefCOCO, RefCOCO+, RefCOCOg, Flickr30K, RefClef, and Ref-reasoning datasets demonstrate the effectiveness of the DGC module and the EGR strategy in consistently boosting the performances of various graph-based REC methods. Without any pretaining, the proposed graph-based method achieves better performance than the state-of-the-art (SOTA) transformer-based methods.
\end{abstract}

% === KEYWORDS ====================================================================
% =================================================================================

% For peer review papers, you can put extra information on the cover
% page as needed:
% \ifCLASSOPTIONpeerreview
% \begin{center} \bfseries EDICS Category: 3-BBND \end{center}
% \fi
%
% For peerreview papers, this IEEEtran command inserts a page break and
% creates the second title. It will be ignored for other modes.
\IEEEpeerreviewmaketitle

% ====================================================================
% ====================================================================
% ====================================================================

% === I. INTRODUCTION =============================================================
% =================================================================================
\section{Introduction}

\IEEEPARstart{R}{eferring} expression comprehension (REC) \cite{qiao2021referring} has become a research hotspot in computer vision and artificial intelligence communities due to its widespread applications ranging from visual surveillance~\cite{Chen_2022_CVPR_visual}, content retrieval~\cite{10.1007/978-3-031-19836-6_10}, visual question answering (VQA)~\cite{Jin2023VQA} to human-computer interaction~\cite{Zhang_2022_CVPR}. 
REC is a text-to-image task in the sense that it aims to locate a specific object in an image specified by a text expression. 
It is still an ongoing and challenging task since it requires understanding the underlying semantics of an expression describing an image, followed by precisely localizing the target object satisfying each condition specified by the expression. 

\begin{figure}[t]
  \centering
  \includegraphics[scale = 0.27]{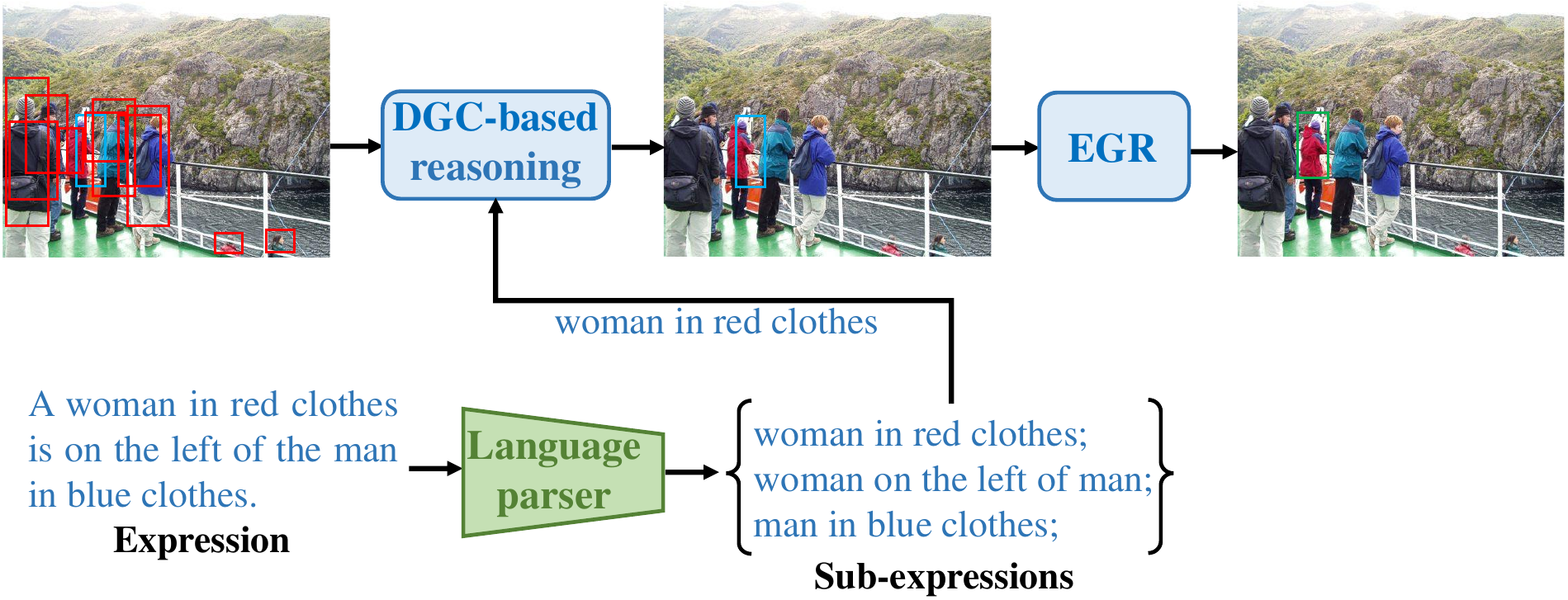}
  %\vspace{-0.1in}
  \caption{
%   Most REC methods only search the object proposals which matches the expression best and take all object proposals for reasoning while some of them are irrelevant to the given expression, thereby degrading reasoning accuracy. 
%   %
%   To address this issue, a language parser is used to convert the expression into several sub-expressions.
  %
  We employ a language parser to decompose a given expression into sub-expressions, each of which is related to a few objects. This helps the DGC module deactivate irrelevant objects during reasoning.
  An expression-guided regression (EGR) strategy is devised to refine the target object location. 
  The red, blue, and green boxes denote the proposals, the proposal best matching the ground truth, and the refined REC output, respectively. 
  %Traditional REC methods usually take all object proposals for reasoning while some of them are unrelated to the given expression and may cause negative impact. In contrast, at each reasoning step, the proposed dynamic gating constraint (DGC) which is guided by relevant sub-expressions can be effectively used to adaptively invalidate unrelated proposals for better message passing and final matching performance.
  }
  \label{fig:teaser}
  %\vspace{-0.15in}
\end{figure} 

Recent top-performing REC algorithms are mainly based on graph-based methods~\cite{DBLP:journals/corr/abs-1812-03426, Liao_2020_CVPR, 10.1145/3394171.3413850, Yang_2019_ICCV, Liu_Wan_Zhu_He_2020, ke2024tmm} and transformer-based methods~\cite{Sun2023referring, Liu_2023_CVPR, Deng_2021_ICCV, Kamath_2021_ICCV}. 
Graph-based methods, in general, implement a two-stage process. 
They first generate the visual features of candidate objects and the textual features of an expression by feeding the image and the expression into an object detector and a language parser, respectively. 
Then, the visual features of each region are used to match the textual features of the expression via multi-step reasoning upon a graph structure. 

By contrast, transformer-based REC typically has a one-stage pipeline. 
The whole image and the given expression are analyzed by a pre-trained convolutional encoder and a language model to extract the visual and textual features, respectively. 
The visual and textual features are fed into the encoder and the decoder of a transformer for joint reasoning and matching. 
Transformer-based methods often achieve better performances than graph-based ones. 
However, the superior performance relies on much higher memory usage and computational costs when conducting the self-attention mechanism.
More importantly, large-scale datasets for model pre-training are required.
% However, their superior performance relies on pretraining on large-scale datasets and complex models. 
% %
% As a result, higher memory usage and computational costs reduce their applicability. 
%
Although graph-based REC methods may not reach comparable performance, they can be trained effectively without any pre-training, and their model sizes are much more compact.

\par
We are inspired by the observation that graph-based REC methods are developed upon off-the-shelf object detectors.
Graph-based REC is quite sensitive to the noise introduced by candidate objects and their connections irrelevant to the expression when performing message passing on graphs for reasoning.
%When conducting reasoning through message passing on graphs, graph-based REC is sensitive to noise caused by candidate objects and their connections that are irrelevant to the expression.

%
Meanwhile, graph-based REC searches for the detected object that best matches the expression. 
However, it results in less accurate REC results when the detector cannot precisely locate the object described by the expression.
In this work, we unleash the potential of graph-based REC by addressing the above two unfavorable issues.

As illustrated in Fig.~\ref{fig:teaser}, we introduce a \textit{plug-and-adapt} module, called dynamic gate constraint (DGC), and an expression-guided regression (EGR) strategy. 
The DGC module is guided by the given expression, and can adaptively determine which graph nodes (candidate objects) and their connections are relevant and should be activated during reasoning for object-text matching. 
Since dynamically deactivating irrelevant nodes and edges is a combinatorial problem and is NP-hard, instead of using the whole expression, we propose to decompose the expression into a set of sub-expressions and gradually apply the DGC module with sub-expressions individually. 
Because each sub-expression involves fewer nodes and connections than the whole expression, this strategy effectively reduces the possibility of involving irrelevant nodes and edges while still covering the whole expression through iterative reasoning. 
The EGR strategy is devised to address the inaccurate detection caused by the detector adopted for graph-based REC. 
We are motivated by the observation that additional textual features inferred from the expression help locate the object.
Thereby, our EGR refines the bounding box prediction by leveraging the information from both the expression and the candidate objects identified after reasoning.

Furthermore, we find that the categorical labels of the detected candidate objects provide additional context information for better reasoning and matching. 
Therefore, we exploit both visual and categorical attention graphs for the REC task. 
Extensive experimental results on six challenging benchmarks demonstrate the effectiveness of the proposed DGC and EGR strategies in processing image-expression pairs with short and long expressions. They both boost the performance of various graph-based REC models to outperform SOTA transformer-based methods without any pre-training.

The main contribution of this work is threefold. 
First, we devise a plug-and-adapt gating module, which is guided by the sub-expressions of a given expression and can adaptively remove irrelevant nodes and connections to the expression describing an image during reasoning, leading to greatly boosted performances of various graph-based REC methods. 
Second, we propose an expression-guided regression strategy that considers both the expression and the selected graph node features to refine the bounding box of an inferred object. 
Expression-guided regression alleviates the issue caused by inaccurate detection of the detector. 
Third, we demonstrate with a comprehensive evaluation on six datasets that, without any pre-training, graph-based REC with the proposed DGC module and EGR strategy performs favorably against the SOTA transformer-based methods.

\section{Related Works}
\label{sec:Related Work}  
In this section, we briefly review the recent development of different REC methods by roughly dividing them into \textit{one-stage} and \textit{two-stage} categories.\\
%. The REC methods can roughly be categorized into one-stage and two-stage methods.
%%
\indent \textbf{One-stage REC Methods:} Generally, one-stage methods first perform feature extraction for the entire image and the expression, followed by cross-modal feature matching to predict the target object. 
The methods in~\cite{Liao_2020_CVPR,Yang_2019_ICCV, huang2021look}  leverage different CNN-based image encoders and language encoders for visual and textual feature extraction, respectively, and then fuse both features (\textit{e.g.}, feature concatenation, inner product) to predict the target object. 
Some transformer-based methods~\cite{ye2022shifting,Yang_2022_CVPR} with different cross-attention modules are proposed to better capture cross-modal feature relationships for the REC task. 
For further performance improvement, the methods in~\cite{wang2022ofa, Kamath_2021_ICCV, Liu_2023_CVPR} leverage pre-trained transformer models for the REC task where the models are first trained using a large and diverse vision and language dataset to learn a universal representation followed by finetuning the models for the target task.\\
\indent \textbf{Two-stage REC Methods:} In contrast, two-stage methods first detect candidate objects from the image, followed by feature extraction from the candidate objects and the expression for matching. 
The method proposed in \cite{Li2018referring} utilizes a visual context LSTM module and a sentence LSTM module to model bundled object context by selecting relevant objects for referring expressions. In this method, all contextual objects are arranged with their spatial locations and progressively fed into the visual context LSTM module to extract and aggregate the context features. 
Hu \textit{et al}.~\cite{Hu_2017_CVPR} propose to decompose the expression into three triples (i.e., subject, preposition/verb, object) and compares them respectively with the object candidates obtained by the detector for matching. 
In~\cite{Chen_Ma_Xiao_Zhang_Chang_2021}, a plug-and-adapt module called Ref-NMS is proposed to remove unrelated candidate objects to the expression through a proposed feature similarity metric. 
In~\cite{Bu2022referring}, besides extracting the visual and textual features, the textual semantics of scene texts in the input image are additionally extracted to align with the referring expressions and visual contents. Such a way of grounding visual representations of expression-correlated scene texts can effectively comprehend scene-text-oriented referring expressions. 

In the two-stage schemes, graph-based methods~\cite{ke2024tmm, Yangsibei_2019_ICCV, 8999516, 10.1145/3548688, 10.1145/3604557, ke2024tmm} have also been widely used for the REC task to capture the relationship between the candidate objects and the expression through multi-step reasoning realized by stacking multiple graph neural network layers. Wang \textit{et al}. introduced two methods~\cite{10.1145/3548688, 10.1145/3604557} for the REC task. One~\cite{10.1145/3548688} involves employing static expression decomposition and directly extracting visual features using off-the-shelf Faster R-CNN, without additional processing for REC reasoning. The other~\cite{10.1145/3604557} applies a residual connections strategy that enables graph-based methods to utilize deeper graph convolution layers, leading to improved performance. Ke~\textit{et al}.~\cite{ke2024tmm} proposed a graph-based method that concurrently constructs a visual graph and a categorical graph based on the objects in a given image and corresponding categories by integrating the CLIP model. Subsequently, guided by the expression, the visual graph and categorical graph collaborate during the reasoning process.
Likewise, Jing \textit{et al}.~\cite{10.1145/3394171.3413902} and Yang \textit{et al}.~\cite{Yang_2020_CVPR} model the relationships between the object candidates and the expression by a semantic graph and a language scene graph, and match the target object in the semantic graph with the guidance of the scene graph.
Chen \textit{et al}.~\cite{Chen_2022_CVPR} propose a graph transformer-based method called M-DGT, which gradually refines the randomly initialized candidate objects to find the target object with the guidance of the expression.\\
\indent With the proposed DGC and EGR modules, our graph-based method outperforms SOTA transformer-based REC methods with much less requirement of computational resources. 
Among the related works, Ref-NMS~\cite{Chen_Ma_Xiao_Zhang_Chang_2021} and M-DGT~\cite{Chen_2022_CVPR} are the closest to our method. 
Different from Ref-NMS, our method employs a language parser to decompose an expression into multiple sub-expressions, each of which is related to fewer candidate objects. 
Gradually utilizing the sub-expressions during the multi-step reasoning stage, the proposed method can better mitigate the influence of irrelevant objects than Ref-NMS. 
M-DGT exploits the candidate-object pruning as a post-processing refinement by only considering the local graph layout after each message-passing step, whereas our method leverages the sub-expressions as the guidance to select the most relevant candidate objects before each message-passing step. 
Thus, M-DGT less effectively reduces the negative influence of irrelevant objects. 
On the other hand, to the best of our knowledge, although the regression strategy for the refinement of inaccurate object bounding boxes is adopted in many transformer-based REC methods, it has not been well-explored for existing graph-based REC methods.

\begin{figure*}[t]
  \centering
  \includegraphics[scale = 0.53]{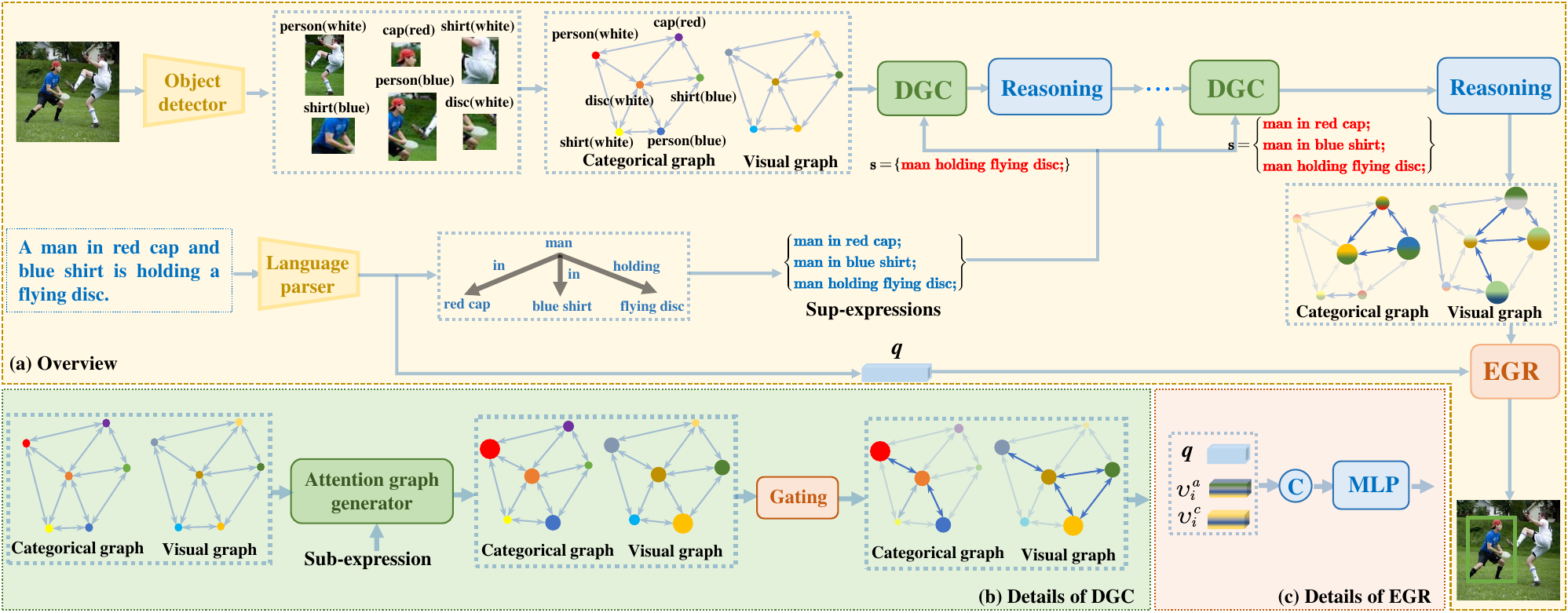}
  %\vspace{-0.05in}
  \caption{
  % yy's version
  Overview of our graph-based REC framework with sub-expressions guided DGC. (a) The flow of activating relevant candidate objects during reasoning adaptively. 
  Bigger nodes in graphs denote larger weights. 
  (b) The DGC module.
  (c) The EGR module, where $\textbf{C}$ represents the concatenation operation.
  Bigger nodes in graphs denote larger weights.  $v_{i}^{a}$ represents the visual feature of the node with the highest score in the visual graph, while $v_{i}^{c}$ represents the corresponding categorical feature in the categorical graph. $\textbf{\textit{q}}$ represents the text feature of the entire expression. 
  Refer main text for more details. 
 }
  \label{fig:framework} 
\end{figure*}

\begin{figure*}[t]
  \centering
  \includegraphics[scale = 0.5]{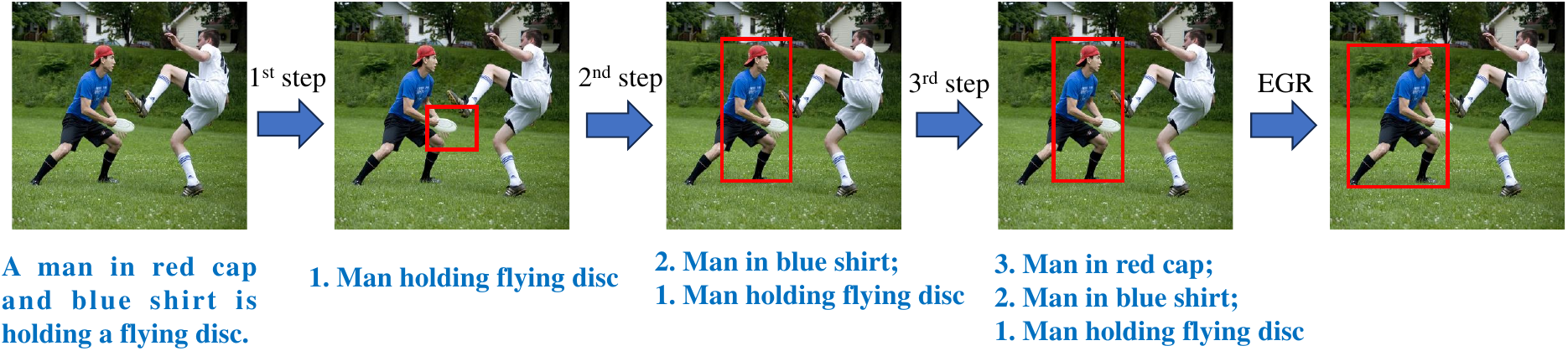}
  %\vspace{-0.05in}
  \caption{Illustration of the reasoning process of our method. As the reasoning steps increase, more sub-expressions are extracted for guidance, following a specific order. In the second and third reasoning steps, our method correctly locates the target object. Finally, we utilize the EGR strategy to refine the predicted bounding box.}
  \label{fig:reasonig} 
\end{figure*}

\section{Proposed Method}
\subsection{Overview of Our Method}\label{sec:framework}

In Fig.~\ref{fig:framework}, (a) illustrates the proposed framework consisting of four modules, including a language parser, a bimodal (visual and categorical) attention graph module, an iterative reasoning module with the dynamic gate constraints (DGC), and an expression-guided regression (EGR) module. (b) and (c) represents the details of the DGC and EGR modules, respectively. 

An input expression is first decomposed into several sub-expressions through the language parser. 
After the detector localizes candidate objects, the bimodal attention graph module constructs the visual and categorical graphs based on the visual features and the categorical labels of the candidate objects, respectively. 
With the dynamic gating mechanism guided by sub-expressions, we iteratively and adaptively deactivate irrelevant objects when passing messages between the two graphs during reasoning. 
The similarity scores between the query expression and the nodes of the two graphs are then evaluated.
The object corresponding to the node with the highest score is finally selected as the target object, where the EGR strategy is further adopted to refine the selected bounding box.

For clarity, we further use the example depicted in Fig~\ref{fig:framework} to show how the proposed DGC module is used to identify the object of interest while filtering out unrelated ones. 
In Fig~\ref{fig:framework}, there are six nodes in the graphs, including person(blue), person(white), cap(red), shirt(blue), shirt(white), and disc(white). 
The expression ``\textit{A man in red cap and blue shirt is holding a flying disc.}'' is parsed into three sub-expressions. 
The proposed DGC module takes these sub-expressions one by one to emphasize the nodes and edges of graphs consistent with the sub-expressions. Thus, the nodes and edges irrelevant to most of the sub-expressions are ignored and then filtered out.
Specifically, in Fig~\ref{fig:reasonig}, the first sub-expression ``\textit{Man holding flying disc}'' is used to guide the DGC module. 
Since the two persons are close to the flying disc, our method currently cannot identify the correct person.
The second sub-expression ``\textit{Man in blue shirt}'' is discriminant.
The DGC module can highlight the nodes and edges that correspond to ``\textit{Man in blue shirt holding flying disc},'' leading to a more accurate localization.
Note that the person in a white shirt is neglected (filtered out) in this sub-expression.
In the third reasoning step, the last sub-expression ``\textit{Man in red cap}'' is combined with the previously processed sub-expressions to form a new sub-expression (i.e., ``\textit{Man in red cap and blue shirt and holding flying disc}''). This sub-expression can guide the DGC module to identify the final set of the related nodes in the graphs. 
Finally, the EGR strategy is utilized to refine the location of the node with the highest score in the visual graph.

In the following, we describe each component of the proposed method in detail.

% Due to the limit of the space, some details of the proposal methods are  in the supplementary materials.
%
%The details are described below.
%-------------------------------------------------------------------------
%\vspace{-10pt}
\subsection{Language Parser}
To better capture the relationship between the expression and the candidate objects in an image, we decompose the expression into several sub-expressions using a language parser where each sub-expression is expected to consist of two nouns (entities) and a preposition or verb between them.
To this end, given an expression $Q$, the expression is parsed into a language scene graph with $T$ sub-expressions by an off-the-shelf language scene graph parser~\cite{Wu_2019_CVPR}.
We denote the language scene graph by ${\cal G}_{\ell}=(V_{\ell}, E_{\ell})$, where $V_{\ell}=\left \{ v_{i}^{\ell}  \right \}_{i=1}^{N}$ and $E_{\ell}=\left \{ e_{ij}^{\ell}  \right \}_{i,j=1}^{N}$ are the sets of the node features and the edge features of the graph, respectively, and $N$ is the number of noun chunks.
As shown in Fig.~\ref{fig:framework}, the nodes and edges of the graph represent the noun chunks and preposition/verb words of the expression, respectively, and their features are encoded by BiLSTM~\cite{burks1954analysis} and GloVe~\cite{pennington2014glove}. 
The features of the whole expression $\mathbf{q}$ are obtained by concatenating the outputs of the last hidden states of both forward and backward BiLSTM.

%-------------------------------------------------------------------------
\subsection{Bimodal Graph Attention Module}
% The previous method~\cite{Yangsibei_2019_ICCV} demonstrates that embedding the textual features of the expression to the graph of an image helps locate the target object described by the expression. We thus construct two graphs.
To better align candidate objects and sub-expressions during reasoning, we leverage both visual appearances and category labels of candidate objects for graph construction since the additional category labels and expressions are in the text domain and have closer semantics than the visual information. 
We thus construct two complete graphs, one for visual features and the other for category labels.  
% and the definitions of the node and edge features are based on the language scene graph ${\cal G}_{\ell}$.
%\subsubsection{Bimodal Graph Construction}

Given an image $I$ with $K$ detected candidate objects and their category labels, we construct a visual graph ${\cal G}_{a}=\left ( V_{a},E_{a},D_{a} \right )$ and a categorical graph ${\cal G}_{c}=\left( V_{c},E_{c},D_{c}\right)$, where $V_{a}=\left \{ v_{i}^{a} \right \}_{i=1}^{K}$, $V_{c}=\left \{ v_{i}^{c} \right \}_{i=1}^{K}$, $E_{a}=\left \{ e_{ij}^{a} \right \}_{i,j=1}^{K}$, $E_{c}=\left \{ e^{c}_{ij} \right \}_{i,j=1}^{K}$, $D_{a}=\left \{ d_{i}^{a} \right \}_{i=1}^{K}$, and $D_{c}=\left \{ d_{i}^{c} \right \}_{i=1}^{K}$ are the node features, the edge features, and the node gating states of ${\cal G}_{a}$ and ${\cal G}_{c}$, respectively. 
Note that we create a node for each candidate object in both graphs.
The details of ${\cal G}_{a}$ and ${\cal G}_{c}$ are described as follows.
% and ,  and  are the node features, edge features and gate statuses of the nodes of ${\cal G}_{c}$, respectively.

% Since we use the GloVe model to embed the labels of $K$ object proposals, we first construct the graph ${\cal G}_{c}$.

%\vspace{-0.2in}
\paragraph{Node Features} For the $i$th node of ${\cal G}_{a}$, its feature is defined by
\begin{equation}
\begin{aligned}
\label{lab:eq3}
&v_{i}^{a} = \mathbf{W}_{a}\left ( \left [\mathbf{o}_{i};\boldsymbol{\mu}_{i}; v_{k_{i}}^{\ell}\right ] \right )+b_{a},
%&c_{i} = \mathbf{W}_{c}\left ( \left [\zeta_{i};\mu_{i}; v_{\ell,k_{i}} \right ] \right )+b_{c},
\end{aligned}
\end{equation}
where $\mathbf{W}_{a}$ is a matrix of trainable parameters and $b_{a}$ is a vector of trainable biases.
$\mathbf{o}_{i}$ is the visual features of the $i$th candidate object. 
$\boldsymbol{\mu}_{i}=\mathbf{W}_{\mu}\left [x_{i},y_{i},w_{i},h_{i},w_{i}h_{i} \right ]$ is the spatial features of the $i$th candidate object, where $\left (x_{i},y_{i} \right )$ is the normalized 2D coordinates of its center. $w_{i}$, $h_{i}$, and $w_{i}h_{i}$ are the normalized width, height, and area, respectively. $\mathbf{W}_{\mu}$ is a trainable matrix. 
$v_{k_{i}}^{\ell}$ is the textual features of the matched noun chunk from the language graph ${\cal G}_{\ell}$, where $k_{i} = \arg \max_{j}\left ( \beta_{i1},...,\beta_{ij}, ..., \beta_{iN} \right )$, and $\beta_{ij}$ is the similarity score between the $j$th node features in graph ${\cal G}_{\ell}$ (i.e., $v_{j}^{\ell}$) and the visual features of the $i$th candidate object $\mathbf{o}_{i}$. $\beta_{ij}$ is computed by
\begin{equation}
\begin{aligned}
\label{lab:eq4}
&\beta_{ij}=\mathbf{W}_{\beta2}\left [ \tanh\left (\mathbf{W}_{\beta1}\mathbf{o}_{i} + \mathbf{W}_{\ell}v_{j}^{\ell}  \right )  \right ],\\
\end{aligned}
\end{equation}
where $\mathbf{W}_{\beta1}$, $\mathbf{W}_{\beta2}$, and $\mathbf{W}_{\ell}$ are trainable matrices. 
Thus, the node features $v_i^a$ in (\ref{lab:eq3}) encodes visual, spatial, and textual information from $\mathbf{o}_{i}$, $\boldsymbol{\mu}_{i}$, and $v_{k_{i}}^{\ell}$, respectively.

\paragraph{Edge Features} The edge features of ${\cal G}_{a}$ are defined as 
\begin{equation}
\label{lab:eq5}
\begin{aligned}
&e_{ij}^{a} = \mathbf{W}_{e}\left ( \left [v_{i}^{a};v_{j}^{a};\boldsymbol{\mu}_{i};\boldsymbol{\mu}_{j} \right ] \right ),\\
\end{aligned}
\end{equation}
where $\mathbf{W}_{e}$ is a trainable matrix of parameters.

\paragraph{Gating States} The gating state of a graph node (\textit{i.e.}, $d_{i}^{a}$ or $d_{i}^{c}$) is either $0$ (``off'') or $1$ (``on''). 
During reasoning, message passing performs only on nodes with the ``on'' state.
The gating states of all nodes of both graphs are initialized as $0$ during graph construction. 
Since the node and edge weights are dynamically computed at each reasoning step, we leave their details in the next subsection.

The node features in (\ref{lab:eq3}), the edge features in (\ref{lab:eq5}), and the node gating state can be compiled for the whole visual graph ${\cal G}_{a}$.
For ${\cal G}_{c}$, its features are computed in a similar way except that the visual features $\mathbf{o}_{i}$ of the candidate objects in (\ref{lab:eq3}) are replaced with the categorical features $\boldsymbol{\zeta}_{i}$, which are obtained by using the GloVe model \cite{pennington2014glove} to embed the object label.

\subsection{Reasoning with Dynamic Gate Constraints}
Once the graphs are constructed, we can proceed to the stage of performing multi-step reasoning like other graph-based methods, \textit{e.g.}~\cite{Yangsibei_2019_ICCV}. At each reasoning step, it is important to identify which candidate objects are irrelevant to the expression since involving them in the message passing at the reasoning phase could introduce severe noise and greatly affect the matching results. 
Therefore, we introduce the DGC module, which is an expression-guided gating mechanism to alleviate the negative impacts caused by those irrelevant objects at each reasoning step.
Before describing the details of gating, we first declare some notations: At the $k$th reasoning step and on each graph ${\cal G}_{*}(k)$ where $* \in \{a, c\}$, $v^{*}_{i}(k)$, $e^{*}_{ij}(k)$, and $d_{i}^{*}(k)$ denote the $i$th node features, the edge features between the $i$th and $j$th nodes, and the gating state of the $i$th node, respectively.
For initialization, we set ${\cal G}_{*}(0)={\cal G}_{*}$, $v^{*}_{i}(0)=v_{i}^{*}$, $e^{*}_{ij}(0)=e^{*}_{ij}$, and $d_{i}^{*}(0)=d_{i}^{*}$. 
At the beginning of the $k$th reasoning step, we initialize all the node weights, edge weights, and gating states to $0$.

\subsubsection{Dynamic Gate Constraint (DGC)}
We only illustrate the details of DGC on ${\cal G}_{a}(k)$, because the DGC on ${\cal G}_{c}(k)$ can be computed similarly.
To determine which nodes should be kept for ${\cal G}_{a}(k)$ for message passing at each reasoning step,
we exploit the correlation between each node and the $k$th sub-expression.
Take Fig.~\ref{fig:framework} with $k=1$ for example.
Since there are two noun chunks in the first sub-expression (\textit{i.e.}, ``man'' holding ``flying disc''), we compute two scores between the $i$th node and the two noun chunks to estimate the correlation extent as follows:
\begin{equation}
\begin{aligned}
\label{lab:eq6}
&\mu_{i}^{a,\delta}=\mathbf{W}_{Q} [ \tanh (v_{i}^{a}(k) + \mathbf{W}_{f}\boldsymbol{\gamma}_{\delta})],\\
\end{aligned}
\end{equation}
where $\mathbf{W}_{f}$ and $\mathbf{W}_{Q}$ are trainable matrices of parameters and $\delta\in\left \{ 1,2 \right \}$. 
$\boldsymbol{\gamma}_{1}$ and $\boldsymbol{\gamma}_{2}$ are the textual features of two noun chunks. The correlation score of the $i$th node is computed by
% \begin{equation}
% \begin{aligned}
% &\tau_{i}^{a}(k)=\max (\mu_{i}^{a,1},\mu_{i}^{a,2}),\\
% \end{aligned}
% \end{equation}
%%
\begin{equation}
\begin{aligned}
\label{lab:eq6_1}
&\tau_{i}^{a}=\max (\mu_{i}^{a,1}, \mu_{i}^{a,2}).\\
\end{aligned}
\end{equation}

If the correlation score of the $i$th node, $\tau_{i}^{a}$, is larger than the average score of all nodes, $\frac{1}{K}{\textstyle \sum_{d=1}^{K}}\tau_{d}^{a}$, it means that the correlation between the $i$th node of ${\cal G}_{a}(k)$ and the $k$th sub-expression is sufficiently strong, and we turn on its gate by setting $d_{i}^{a}(k)$ to $1$.
This process is repeated for each node on every graph. It is essential to note that if an expression is too short to be decomposed (\textit{i.e.}, comprising only one or two noun chunks), the expression is directly fed into the sup-expression set $S$. In the case of comprising only a single noun chunk, we set $\boldsymbol{\gamma}_{1}=\boldsymbol{\gamma}_{2}$. 

\subsubsection{Reasoning on Bimodal Sub-Graphs}
After applying DGC, we get two corresponding sub-graphs ${\cal G}_{a}^\mathrm{sub}$ and ${\cal G}_{c}^\mathrm{sub}$ consisting of the active nodes and their associated edges from ${\cal G}_{a}(k)$ and ${\cal G}_{c}(k)$, respectively, at the $k$th reasoning step. 
We describe the details of ${\cal G}_{a}^\mathrm{sub}$, and those of ${\cal G}_{c}^\mathrm{sub}$ are similar except the visual features are replaced with the categorical features. 
The sets of node and edge features of ${\cal G}_{a}^\mathrm{sub}$ are denoted as $V_{a}^\mathrm{sub}$ and $E_{a}^\mathrm{sub}$, respectively, where $V_{a}^\mathrm{sub} = \{v_{i}^{a}(k)| d_{i}^{a}(k)=d_{i}^{c}(k)=1, \ i=1,...,K\}$, $E_{a}^\mathrm{sub} = \{e_{ij}^{a}(k)|v_{i}^{a}(k) \in {\cal G}_{a}^\mathrm{sub}\ \mbox{and}\  v_{j}^{a}(k) \in {\cal G}_{a}^\mathrm{sub},  \ i,j=1,...,K\}$, and $\left |V_{a}^\mathrm{sub} \right |=M$ is the number of the nodes in ${\cal G}_{a}^\mathrm{sub}$. Note that $\left | V_{a}^\mathrm{sub} \right |=\left | V_{c}^\mathrm{sub} \right |=M$. If $V_{a}^\mathrm{sub}$ is empty, we construct an alternative $V_{a}^{sub}$ by relaxing the gate constraints. In particular, we first create a sub-graph $V_{c}^{sub}$ by using the gate states ($d_{i}^{c}(k)=1$) in ${\cal G}_{c}(k)$. Then, we leverage $V_{c}^{sub}$ to find the corresponding nodes and edges from visual graph ${\cal G}_{a}(k)$ to establish $V_{a}^{sub}$.
%
%without loss of generality, we assume the remaining nodes in $V_{a}^{sub}$ are the first $M$ nodes in ${\cal G}_{c}^{sub}$ for the simplicity of indexing.

% The sets of node and edge features of ${\cal G}_{a}^{sub}$ are denoted as $V_{a}^\mathrm{sub}=\left \{ v_{i}^{a,\mathrm{sub}} \right \}_{i=1}^{M}$ and $E_{a}^{sub}=\left \{ e_{ij}^{a,\mathrm{sub}} \right \}_{i,j=1}^{M}$ respectively, where $\left | V_{a}^{sub} \right |=M$, $V_{a}^{sub} = \{v_{i}^{a}|v_{i}^{a} \in {\cal G}_{a}(k), d_{i}^{a}(k)=1\ \mbox{and} \ d_{i}^{c}(k)=1\}$, and $E_{a}^{sub} = \{e_{ij}^{a}|v_{i}^{a} \in {\cal G}_{a}^{sub}(k)\ \mbox{and}\  v_{j}^{a} \in {\cal G}_{a}^{sub}(k) \}$. 

% For $E_{a}^{sub}$, if $k=0$, $E_{a}^{sub}$ is the subset of $E_{a}$ based on $V_{a}^{sub}$. Otherwise, the definition of $e_{ij}^{a,sub}$ is similar as eq. \eqref{lab:eq5}. 
%%
%\vspace{-0.15in}
\paragraph{Node Weights:} We compute the normalized node weights on the visual sub-graph ${\cal G}_{a}^\mathrm{sub}$ via
\begin{equation}
\begin{aligned}
&w_{i}^{a,\mathrm{sub}}=\frac{\exp (\tau_{i}^{a,\mathrm{sub}} )}{\sum_{j=1}^{M}\exp (\tau_{j}^{a,\mathrm{sub}} )},\\
\end{aligned}
\end{equation}
\noindent where $\tau_{i}^{a,\mathrm{sub}}$ is the corresponding correlation score for the $i$th node of ${\cal G}_{a}^\mathrm{sub}$ as computed in \eqref{lab:eq6_1}. 

%\vspace{-0.15in}
\paragraph{Edge Weights:} For the edge weights of ${\cal G}_{a}^\mathrm{sub}$, we first compute the non-normalized edge score between the $i$th and $j$th nodes of ${\cal G}_{a}^\mathrm{sub}$ by
%$v_{i}^{a,\mathrm{sub}}$ and $v_{j}^{a,sub}$ by
\begin{equation}
\begin{aligned}
&\mu_{ij}^{a,\mathrm{sub}}=\mathbf{W}_{\nu} [ \tanh (e_{ij}^{a,\mathrm{sub}}(k) + \mathbf{W}_{s}f_{s}(k))],\\
\end{aligned}
\end{equation}
\noindent where $\mathbf{W}_{\nu}$ and $\mathbf{W}_{s}$ are trainable matrices of parameters, $f_{s}(k)$ is the concatenated textual features of all the sub-expressions in the set $S$ which consists of all the visited sub-expressions from the beginning to the $k$th reasoning step. If $v_{i}^{a,\mathrm{sub}}$ and $v_{j}^{a,\mathrm{sub}}$ are strongly correlated with the noun chunks in the sub-expression, the edge score between these two nodes should be high. Therefore, similar to the process of determining the node gating state, we only consider the connected edges to the $i$th node of ${\cal G}_{a}^\mathrm{sub}$ whose edge scores $\mu_{ij}^{a,\mathrm{sub}}$ are larger than the threshold $\xi=\frac{1}{M}{\textstyle \sum_{j=1}^{M}}\mu_{ij}^{a,\mathrm{sub}}$. The final edge weights are computed as follows: 
\begin{equation}
\begin{aligned}
w_{ij}^{a,\mathrm{sub}}=\frac{\exp\left( \mu_{ij}^{a,\mathrm{sub}}  \right)  \cdot\mathbbm{1}[\mu_{ij}^{a,\mathrm{sub}} > \xi]}{\sum_{m=1}^{\lambda_{i}}\exp\left( \mu_{im}^{a,\mathrm{sub}} \right)  \cdot\mathbbm{1}[\mu_{im}^{a,\mathrm{sub}} > \xi]},\\
\end{aligned}
\end{equation}
\noindent where $\lambda_{i}$ is the number of the edges connected to the $i$th node of ${\cal G}_{a}^\mathrm{sub}$ and $\mathbbm{1}(\cdot)$ is the indicator function.

% The definition of node features, edge features, node weights and edge weights of ${\cal G}_{c}^{sub}$ are similar to ${\cal G}_{a}^{sub}$ as above,  Note that $\left | V_{c}^{sub} \right |=\left | V_{a}^{sub} \right |=M$.

Once the node and edge weights have been computed, we follow the same message passing computation upon both sub-graphs respectively as ~\cite{Yangsibei_2019_ICCV} to complete the $k$th reasoning step where it is equivalent that we update the node features of the corresponding nodes with the ``on'' state in ${\cal G}_{a}(k)$ and ${\cal G}_{c}(k)$ for the next reasoning step. Then, we repeat the same reasoning process until all the sub-expressions in the language graph have been visited. Specifically, the $i$th node feature $v_i^a(k)$ of ${\cal G}_{a}(k)$ whose gate state is ``on'' (\textit{i.e.}, $d_i^a(k)=1$) can be computed as follows:
\begin{equation}
\begin{aligned}
&v_{i}^{a}(k)=\mathbf{W}_{k}\left (\widetilde{v}_{i}^{a}(k)+ \widehat{v}_{i}^{a}(k)\right ) + v_{i}^{a}(k-1), \\
\end{aligned}
\end{equation}
where $\mathbf{W}_{k}$ is a trainable matrix of parameters, and $\widetilde{v}_{i}^{a}(k)$ and $\widehat{v}_{i}^{a}(k)$ are the aggregated features of the node features from the active connected neighboring nodes of the $i$th node and the transformed feature for the self-loop of the $i$th node at the $(k-1)$th reasoning step, respectively. More Specifically,
\begin{equation}
\begin{aligned}
&\widetilde{v}_{i}^{a}(k)=\sum_{\tiny
\begin{array}{l}
w_{ij}^{a}>0,\\
d_{j}^{a}(k)=1
\end{array}
}w_{ij}^{a}\left ( \widetilde{\mathbf{W}}_{k} v_{j}^{a}(k-1) w_{j}^{a}
+ \widetilde{b}_{k} \right ),\\
&\widehat{v}_{i}^{a}(k)=w_{i}^{a}\widehat{\mathbf{W}}_{k}v_{i}^{a}(k-1) +\widehat{b}_{k},\\
\end{aligned}
\end{equation}
where $\widetilde{\mathbf{W}}_{k}$ and $\widehat{\mathbf{W}}_{k}$ are trainable matrices of parameters, and $\widetilde{b}_{k}$ and $\widehat{b}_{k}$ are trainable biases. when $k-1=0$, $v_{i}^{a}(k-1)=v_{i}^{a}$.

The computation of message passing for ${\cal G}_{c}(k)$ is similar to ${\cal G}_{a}(k)$, except the superscript ``$a$'' in the above-mentioned equations are replaced with ``$c$''.

\begin{table*}[t]
\tiny
\caption{Comparisons among our method and other existing methods on the RefCOCO, RefCOCO+, RefCOCOg, Flickr30k, Refelf, and Ref-reasoning datasets when the detected bounding boxes are used. The first, second, third, and fourth groups of the table are transformer-based methods, non-transformer-based methods, and graph-based methods with DGC and our method, respectively. In addition, CC~\cite{sharma-etal-2018-conceptual}, SBU~\cite{ordonez2011im2text}, COCO~\cite{Pont-Tuset_2015_ICCV}, VG~\cite{krishna2017visual}, Flickr30K~\cite{Plummer_2015_ICCV}, OI~\cite{Plummer_2015_ICCV}, and Objects365V1~\cite{Shao_2019_ICCV} are the datasets used for pre-training. The best and the second-best results are marked in bold and underlined, respectively. CO, F30K, O365V1 represent COCO, Flickr30K and Objects365V1, respectively.}
%\vspace{-0.05in}
\label{tab:total1}
\centering
\renewcommand\arraystretch{1}{
\setlength{\tabcolsep}{0.8mm}{
\resizebox{1\textwidth}{!}{%
\begin{tabular}{ccccccccccccc}
\hline
% \toprule
\multirow{2}{*}{Methods}
&\multirow{2}{*}{Pre-train Dataset}
&\multicolumn{3}{c}{RefCOCO} 
&\multicolumn{3}{c}{RefCOCO+} 
&\multicolumn{2}{c}{RefCOCOg} & Flickr30k& Refclef & Ref-reasoning\\
\cline{3-13}
 & &val&testA&testB&val&testA&testB&val&test &test &test &test\\
\hline
%VGTR~\cite{9859880}&None&79.30&82.32&73.78&64.40&70.09&56.51&66.83&67.28&-&-&-\\
TransVG~\cite{Deng_2021_ICCV}&CO&81.02&82.72&78.35&64.82&70.70&59.64&68.67&67.73&79.10&70.73&-\\
M-DGT~\cite{Chen_2022_CVPR}&None&85.37&83.01&\uline{85.24}&70.02&72.26&68.92&79.21&79.06&79.97&-&-\\
VLTVG~\cite{Yang_2022_CVPR}&CO&84.77&87.24&80.49&74.19&78.93&65.17&76.04&74.18&79.84&71.98&-\\
QRNet~\cite{ye2022shifting}&CO&84.01&85.85&82.34&72.94&76.17&63.81&73.03&72.52&81.95&74.61&-\\
TransVG++~\cite{deng2022transvg++} &CO&86.28&88.37&80.97&75.39&80.45&66.28&76.18&76.30&81.49&74.70&-\\
SeqTR~\cite{zhu2022seqtr} &VG&83.72&86.51&81.24&71.45&64.88&66.28&74.86&74.21&-&-&-\\
%PEVL~\cite{arxiv.2205.11169} &CC, SBU, COCO, VG&89.60&92.50&85.00&83.00&88.40&74.50&87.10&86.30\\
VinVL-L~\cite{Zhang_2021_CVPR} &CO, OI, O365V1, VG&81.80&87.20&74.30&74.50&80.80&64.30&74.60&75.70&-&-&-\\
%UNICORN~\cite{yang2021crossing} &COCO, VG, Flickr30K&88.30&90.40&83.10&80.30&85.10&71.90&83.40&83.90\\
%UniTAB~\cite{DBLP:journals/corr/abs-2111-12085} &COCO, VG, Flickr30k&88.59&91.06&83.75&80.97&85.36&71.55&84.58&84.70\\

%VL-BERT-L~\cite{su2019vl}&CC&-&-&-&72.34&78.52&62.61&-&-&-&-&-\\
%UNITER-L \cite{10.1007/978-3-030-58577-8_7} &CC, SBU, CO, VG&81.41&87.04&74.17&75.90&81.45&66.70&74.86&75.77&-&-&-\\
%VILLA-L \cite{gan2020large} &CC, SBU, CO, VG&82.39&87.48&74.84&76.17&81.54&66.84&76.18&76.71&-&-&-\\
ERNIE-ViL L~\cite{Yu_Tang_Yin_Sun_Tian_Wu_Wang_2021}&CC, SBU&-&-&-&75.95&82.07&66.88&-&-&-&-&-\\
%ViLBERT~\cite{lu2019vilbert}&CC&-&-&-&72.34&78.52&62.61&-&-&-& -&-\\
MDETR~\cite{Kamath_2021_ICCV} &CO, VG, F30K&86.57&89.58&81.41&79.52&84.09&70.62&81.64&80.89&83.80&-&-\\
OFA-base~\cite{wang2022ofa} &VG&\uline{88.48}&\textbf{90.67}&83.30&\uline{81.39}&\uline{87.15}&\textbf{74.29}&\uline{82.29}&\uline{82.31}&-&-&-\\
AMC~\cite{Yang_2023_CVPR} &CO, SBU, CC, VG&-&-&-&-&-&-&-&-&\uline{86.59}&-&-\\
VG-LAW~\cite{Su_2023_CVPR}&CO&86.62&89.32&83.16&76.37&81.04&67.50&76.90&76.96&-&\textbf{77.22}&-\\
SMCIM~\cite{10345481}&CO&85.10&88.23&80.08&74.44&79.48&665.21&77.25&75.78&-&75.18&-\\
LGR-NET~\cite{10463072}&CO&85.63&88.24&82.69&75.32&80.60&68.30&76.82&77.03&81.97&74.64&-\\
PFOS~\cite{9699024}&CO&78.44&81.49&73.13&65.86&72.43&55.26&67.89&67.63&-&67.90&-\\
\hline
RCCF~\cite{Liao_2020_CVPR}&None&-&81.06&71.85&-&70.35&56.32&- &65.73&-&63.79&-\\
%FAOA~\cite{Yang_2019_ICCV}&None&-&-&-&-&-&-&-&-&68.71&60.67&-\\
%MCN~\cite{luo2020multi}&None&78.93&82.29&74.98&65.54&61.11&50.93&65.76&61.56\\
%LBYL~\cite{huang2021look}&None&79.67&82.91&74.15&68.64&73.38&59.49&-&-&-&66.51&-\\
LBYL-RED~\cite{Huang_Qin_Qi_Sun_Zhang_2022}&None&80.97&83.20&77.66&69.48&73.80&62.20&71.11&70.67&-&67.27&-\\
%MattNet~\cite{yu2018mattnet}&None&76.65&80.43&69.28&65.33&70.26&56.00&66.58&67.01\\
%NMT~\cite{liu2019learning}&None&76.41&81.21&70.09&66.46&72.02&57.52&65.87&66.44\\
%CM-Att~\cite{liu2019improving}&None&78.35&83.14&71.32&68.09&73.65&58.03&67.99&68.67\\
% \multicolumn{1}{|c|}{MMI \cite{Mao_2016_CVPR}}&None&64.90&54.51&54.03&42.81&-\\
% \multicolumn{1}{|c|}{CG \cite{Luo_2017_CVPR}}&None&67.94&55.18&57.05&43.33&-\\
VSGM~\cite{10.1145/3394171.3413902}&None&82.66&82.06&84.24&67.70&69.34&65.74&75.73&76.87&-&-&-\\
%LGRAN~\cite{Wang_2019_CVPR}&None&-&76.60&66.40&-&64.00&53.40&- &62.50&-&-&-\\
Ref-NMS~\cite{Chen_Ma_Xiao_Zhang_Chang_2021}&None&80.70&84.00&76.04&68.25&73.68&59.42&70.55&70.62& & &\\
DGA~\cite{Yangsibei_2019_ICCV}&None&-&78.42&65.53&-&69.07&51.99&-&63.28&-&-&45.87\\
CMRIN~\cite{8999516}&None&-&82.53&68.58&-&75.76&57.27&-&67.38&-&-&45.87\\
SGMN~\cite{Yang_2020_CVPR}&None&-&82.08&71.75&-&74.52&56.83&-&68.80&-&-&51.39\\
% \hline
% \multicolumn{10}{c}{\textbf{Transformer-based method}}\\

% \hline

% \multicolumn{10}{c}{\textbf{Graph-based method w/ DGC}}\\
% \hline
% DGA w/ DGC &None&-&81.14&80.96&-&77.98&53.37&-&65.74\\
% CMRIN w/ DGC &None&-&85.38&81.14&-&87.75&60.08&-&67.87\\
% SGMN w/ DGC&None&-&86.13&80.82&-&88.06&61.27&-&68.55\\
%\multicolumn{1}{c}{VSGM+DGC}&None&86.13&80.82&88.06&61.27&68.55\\
% \hline
% \multicolumn{10}{c}{\textbf{Ours}}\\
ECMGANs~\cite{10.1145/3548688}&None&87.05&87.98&86.03&75.52&80.62&70.43&80.76&80.79&-&-&-\\
LRGAT-VG+LGDA~\cite{10.1145/3604557}&None&87.68&88.55&87.49&76.56&80.67&71.08&82.15&81.80&-&-&-\\
CLIPREC~\cite{ke2024tmm}&None&-&84.63&84.51&-&76.82&63.07&-&76.83&-&-&-\\
\hline
%CGN &None&82.58&83.96&83.58&70.08&74.56&62.54&72.23&73.68\\
DGA w/ DGC &None&-&81.14&80.96&-&77.98&59.75&-&65.74&-&-&50.52\\
CMRIN w/ DGC &None&-&85.38&81.14&-&79.75&60.08&-&67.87&-&-&49.84\\
SGMN w/ DGC&None&-&85.13&80.82&-&80.06&61.27&-&68.55&-&-&\uline{55.67}\\
\hline
%Our network w/ DGC&None&85.63&87.82&85.35&75.65&83.58&65.03&77.63&78.09& & &\\
Ours &None&\textbf{88.72}
&\uline{89.93}&\textbf{87.85}&\textbf{82.22}&\textbf{88.16}&\uline{73.02}&\textbf{83.27}&\textbf{83.36}&\textbf{86.83}&\uline{75.75}&\textbf{62.64}\\
\hline
\end{tabular}}}
}
%\vspace{-5pt}
\end{table*}

\begin{table}[t]
\caption{Study on different detectors. C, VG, and VOC represent Faster R-CNN trained on COCO, VG, and VOC, respectively. m@0.5 represents mAP@0.5.
%Ablation results of our method under two strategies of processing sub-expressions.
}
\label{tab:ablationde}
\centering
\renewcommand\arraystretch{1}{
\setlength{\tabcolsep}{1.5mm}{
\resizebox{0.46\textwidth}{!}{
\begin{tabular}{ccccccccc}
\hline
% \toprule
\multirow{2}{*}{VOC}
&\multirow{2}{*}{VG}
&\multirow{2}{*}{C}
&\multirow{2}{*}{m@0.5}
&\multicolumn{2}{c}{RefCOCO} 
&\multicolumn{2}{c}{RefCOCO+} 
&\multicolumn{1}{c}{RefCOCOg}
\\
\cline{5-9}
\multicolumn{1}{c}{}
&\multicolumn{1}{c}{}
& & &testA&testB&testA&testB&test\\
\hline
\checkmark & & &75.86&83.85&82.67&85.52&69.58&79.96\\
&\checkmark
& &81.93&86.67&85.90&86.03&71.82&82.28\\
\checkmark &\checkmark & \checkmark &87.24&88.73&87.08&87.24&71.67&81.28\\
& & \checkmark &\textbf{88.72}&\textbf{89.93}&\textbf{87.85 }&\textbf{88.16}&\textbf{73.02}&\textbf{83.36}\\
\hline
\end{tabular}}}}
%\vspace{-5pt}
\end{table}
% %%
% \subsection{Matching and Expression-guided Regression}
% % Once all the sub-expressions in $S$ have been processed at the $N$th reasoning step, we design two losses to combine the final sub-graphs and the whole expression. Since we hope the ground-truth node include more information of the expression, we try to make the ground-truth node of both sup-graphs and the expression as similar as possible.

% %%
% % In other words, we are trying to project the visual features and GloVe features into a common subspace for matching. 
% %%
\subsection{Matching and Expression-guided Regression}
Once the reasoning is completed after $T$ reasoning steps (\textit{i.e.}, $T$ is the total number of sub-expressions in ${\cal G}_{\ell}$), for the matching loss, we first compute the similarity scores between the final aggregated features of each node of two graphs (i.e., ${\cal G}_{a}(T)$ and ${\cal G}_{c}(T)$ ) and the textual features of the whole expression (i.e., $\mathbf{q}$) as follows:
\begin{equation}
\begin{aligned}
&\vartheta_{i}^{*}= \left \langle  \frac{\mathbf{W}_{v*}v_{i}^{*}(T)}{\| \mathbf{W}_{v*}v_{i}^{*}(T) \|},  \frac{\mathbf{W}_{q}\mathbf{q}}{\| \mathbf{W}_{q}\mathbf{q} \| } \right \rangle,\\
\end{aligned}
\end{equation}
where $* \in \{a, c\}$ and $\mathbf{W}_{v*}$ and $\mathbf{W}_{q}$ are trainable matrices of parameters. 
%

%
% The score of the ground-truth node of $G_{\rho}^{(N)}$ (GT) is denoted as $score_{gt}^{\rho}$. 
The scores of the ground-truth (GT) nodes of both graphs are denoted as $\vartheta_\mathrm{GT}^{*}$.
We then compute the loss to bridge the two graphs as follows:
% upon the final active nodes after the last reasoning step as follows:
\begin{equation}
\begin{aligned}
&\mathcal{L}_\mathrm{CE} = -\log \left ( P_\mathrm{GT}\right ),
\end{aligned}
\end{equation}
\begin{equation}
\begin{aligned}
&P_\mathrm{GT}=\frac{\exp\left ( \vartheta_\mathrm{GT}^{a}+\vartheta_\mathrm{GT}^{c} \right )}{\sum_{i=1}^{K}\exp\left (\vartheta_{i}^{a}+\vartheta_{i}^{c}\right)}, 
\end{aligned}
\end{equation}
where $K$ is the number of nodes of both graphs. 

To refine the bounding box predictions, we first concatenate the features of the entire expression and two ground-truth nodes in both graphs. Then, the concatenated features are used to refine the bounding box of the target object by adding additional fully-connected layers and enforcing the smooth $\mathcal{L}_1$ loss, $\mathbf{SmoothL}_{1}(\cdot,\cdot)$, between the predicted and ground-truth bounding boxes as follows: 
% Finally, we compute similarity between the repredicted bounding box and ground-truth bounding box. The details of our designed regression loss are showed as follow:
\begin{equation}
\begin{aligned}
&b_\mathrm{pred}=\mathbf{MLP}\left ( \left [ v_\mathrm{GT}^{a}(T);v_\mathrm{GT}^{c}(T)\right ];\mathbf{q} \right ),\\
&\mathcal{L}_\mathrm{reg}=\mathbf{SmoothL}_{1}(b_\mathrm{pred},b_\mathrm{GT}),\\
\end{aligned}
\end{equation}
where $b_\mathrm{pred}$ is the bounding box expressed as a 4-D vector predicted by the multi-layer perception function $\mathbf{MLP}(\cdot)$ with learned parameter set $\mathbf{q}$ and $b_\mathrm{GT}$ is the ground-truth bounding box. So, the final loss of our method is defined as
\begin{equation}
\begin{aligned}
&\mathcal{L}=\mathcal{L}_\mathrm{CE}+\mathcal{L}_\textrm{reg}.\\
\end{aligned}
\end{equation}

During inference, if the Intersection-over-Union (IoU) value between the refined bounding box of the node with the highest score (\textit{i.e.}, $\vartheta_{i}^{a}+\vartheta_{i}^{c}$) and the ground-truth bounding box is larger than $0.5$, we consider the test sample is predicted correctly.

\begin{table*}[t]
\caption{Ablation studies of our method when only the DGC module or the EGR strategy is used. D and E represent DGC and EGR, respectively.
%Ablation results of our method under two strategies of processing sub-expressions.
}
%\vspace{-0.05in}
\label{tab:ablation3}
\centering
\renewcommand\arraystretch{1.2}{
\resizebox{0.9\textwidth}{!}{%
\setlength{\tabcolsep}{5mm}{
\begin{tabular}{cccccccccc}
\hline
% \toprule
\multirow{2}{*}{D}
&\multirow{2}{*}{E}
&\multirow{2}{*}{D+E}
&\multicolumn{2}{c}{RefCOCO} 
&\multicolumn{2}{c}{RefCOCO+} 
&\multicolumn{1}{c}{RefCOCOg}
&\multicolumn{1}{c}{Ref-reasoning}\\
\cline{4-9}
\multicolumn{1}{c}{}
&\multicolumn{1}{c}{}
& &testA&testB&testA&testB&test&test\\
\hline
\multicolumn{1}{c}{\ding{55}} %\usym{2718} \ding{56}
&\multicolumn{1}{c}{\ding{55}}
&\ding{55}&83.96&83.58&74.56&62.54&73.68&51.63\\
\multicolumn{1}{c}{\checkmark}
&\multicolumn{1}{c}{}
& &87.82&85.35&83.58&65.03&78.09&58.76\\
\multicolumn{1}{c}{} 
&\multicolumn{1}{c}{\checkmark}
& &86.67&85.83&76.28&65.35&75.56&55.65\\
\multicolumn{1}{c}{} 
&\multicolumn{1}{c}{}
& \checkmark&\textbf{89.93}&\textbf{87.85}&\textbf{88.16}&\textbf{73.02}&\textbf{83.36}&\textbf{62.64}\\
\hline
\end{tabular}}}}
%\vspace{-15pt}
\end{table*}

\begin{table}[!t]
\caption{Accuracy rates (\%) of UniVSE on the manually-labeled expressions selected from the Ref-reasoning and RefCOCOg datasets.}
\label{tab:parser}
\centering
\renewcommand\arraystretch{1.2}{
\resizebox{0.3\textwidth}{!}{%
\begin{tabular}{ccc}
\hline
\toprule
Parser &Ref-reasoning & RefCOCOg
\\
\cline{1-3}
UniVSE &96.67&97.83\\
\hline
\end{tabular}}}
\end{table}

\begin{table*}[t]
\caption{Ablation studies on the visual graph ${\cal G}_{a}$, the categorical graph ${\cal G}_{c}$, and the ${\cal G}_{a}$+${\cal G}_{c}$ constraint. 
%Ablation results of our method under two strategies of processing sub-expressions.
}
%\vspace{-0.05in}
\label{tab:ablation2}
\centering
\renewcommand\arraystretch{1.2}{
\setlength{\tabcolsep}{5mm}{
\resizebox{0.8\textwidth}{!}{
\begin{tabular}{cccccccc}
\hline
% \toprule
\multirow{2}{*}{${\cal G}_{a}$}
&\multirow{2}{*}{${\cal G}_{c}$}
&\multirow{2}{*}{${\cal G}_{a}$+${\cal G}_{c}$}
&\multicolumn{2}{c}{RefCOCO} 
&\multicolumn{2}{c}{RefCOCO+} 
&\multicolumn{1}{c}{RefCOCOg}
\\
\cline{4-8}
\multicolumn{1}{c}{}
&\multicolumn{1}{c}{}
& &testA&testB&testA&testB&test\\
\hline
\multicolumn{1}{c}{\checkmark}
&\multicolumn{1}{c}{}
& &87.72&86.07&87.23&71.68&82.09\\
\multicolumn{1}{c}{} 
&\multicolumn{1}{c}{\checkmark}
& &84.65&83.28&84.06&68.63&78.96\\
\multicolumn{1}{c}{} 
&\multicolumn{1}{c}{}
& \checkmark&\textbf{89.93}&\textbf{87.85}&\textbf{88.16}&\textbf{73.02}&\textbf{83.36}\\
\hline
\end{tabular}}}}
%\vspace{-5pt}
\end{table*}

\section{Experimental Results}
\label{sec:experiment}

\subsection{Datasets}
We present extensive evaluation results for our method on six challenging REC benchmarks~\cite{qiao2021referring}: RefCOCO~\cite{kazemzadeh-etal-2014-referitgame}, RefCOCO+~\cite{kazemzadeh-etal-2014-referitgame}, RefCOCOg~\cite{Mao_2016_CVPR}, Flickr30K entities~\cite{Plummer_2015_ICCV}, RefClef~\cite{kazemzadeh-etal-2014-referitgame}, and Ref-reasoning~\cite{Yang_2020_CVPR}, where RefCOCO and RefCOCO+ has expressions with an average length of $3.61$ 
 and $3.65$ words, respectively. In contrast, RefCOCOg and Ref-reasoning contain longer expressions, with averages of $8.4$ and $8.5$ words, respectively. Moreover, the proportion of short expressions (\textit{i.e.}, containing only one or two noun chunks) in RefCOCO, RefCOCO+, RefCOCOg, and Ref-reasoning is $85.26\%$, $87.73\%$, $46.77\%$, and $19.68\%$, respectively. We briefly describe the details of the datasets used in our work for training and evaluations below.

\begin{itemize}
 \item \textbf{RefCOCO}, \textbf{RefCOCO+}, and \textbf{RefCOCOg:} RefCOCO and RefCOCO+ contain $142,210$ and $141,564$ expressions for $50,000$ and $49,856$ objects in $19,994$ and $19,992$ images, respectively. In particular,~\cite{kazemzadeh-etal-2014-referitgame}, RefCOCO is split into the training, validation, testA, and testB sets with $120,624$, $10,834$, $5,657$ and $5,095$ expression-referent pairs, respectively, where testA mainly consists of the images with multiple people while testB includes images with multiple other objects. Besides, RefCOCO+ is also organized in the same four-split setting with $120,191$, $10,758$, $5,726$, and $4,889$ expression-referent pairs for the training, validation, testA, and testB sets, respectively. Unlike RefCOCO, the expressions of RefCOCO+ do not contain descriptions of absolute location. RefCOCOg contains 95,010 long expressions for $49,822$ objects in $25,799$ images with $80,512$, $4,896$, and $9,602$ expression-referent pairs for the training, validation, and test sets, respectively. RefCOCO, RefCOCO+, and RefCOCOg are curated from the MSCOCO dataset~\cite{Pont-Tuset_2015_ICCV} with $80$ object categories.
\item \textbf{Flickr30K:} It contains $31,783$ images, each described in five sentences. For each sentence, the corresponding referred objects are from one of seven common object categories and an ``other'' category for objects that do not belong to any of the other seven object categories. The numbers of training, validation, and test samples of this dataset are $425,831$, $14,433$, and $14,481$, respectively.
\item \textbf{RefClef:} It contains $19,997$ images selected from the SAIAPR-12 dataset with $54,127$, $5,842$, and $60,103$ referring expressions for the training, validation, and test sets, respectively.
\item \textbf{Ref-reasoning:} It is collected from the GQA dataset~\cite{hudson2019gqa} that it contains $108,077$ images where the numbers of referring expressions in training, validation, and test sets are $721,164$, $36,183$, and $34,609$, respectively.
\end{itemize}

\subsection{Implementation Details}
For fair comparisons with existing SOTAs, we in this work adopt the object detector Faster R-CNN~\cite{Anderson_2018_CVPR} pre-trained on the COCO dataset to obtain the features and the predicted labels of the detected objects where the backbone of the region proposal network (RPN) of the detector is ResNet-101. Additionally, we use an open-sourced color detector\footnote{\url{https://github.com/stefanbo92/color-detector}} based on OpenCV library to derive the color attribute once the detected objects have been obtained. The combination of the color attributes and object labels form the categorical graph. During training, we remove the overlapping image categories between the training sets of COCO and the six REC datasets used in our paper. Although the setting is similar to MATTNet~\cite{yu2018mattnet}, ours is more restricted since we remove the overlapping images from three more REC datasets.
There are two standard REC settings for evaluation: one that uses both the ground truth and detected object information for graph construction and the other that only uses detected objects.
For the purpose of fair comparisons with the transformer-based methods, we show the results of the latter setting where we adopt accuracy as the evaluation metric, and a prediction is considered correct if the IoU value between the predicted bounding box of the detected object and ground-truth bounding box is greater than $0.5$. 
The number of epochs, batch size, and learning rate for training are set to $30$, $32$, and $10^{-4}$, respectively. The optimizer we utilized is Adam with $\beta_{1}$ set to $0.8$ and $\beta_{2}$ set to $0.9$.

\begin{figure}[t]
  \centering
  \includegraphics[width = 0.4\textwidth]{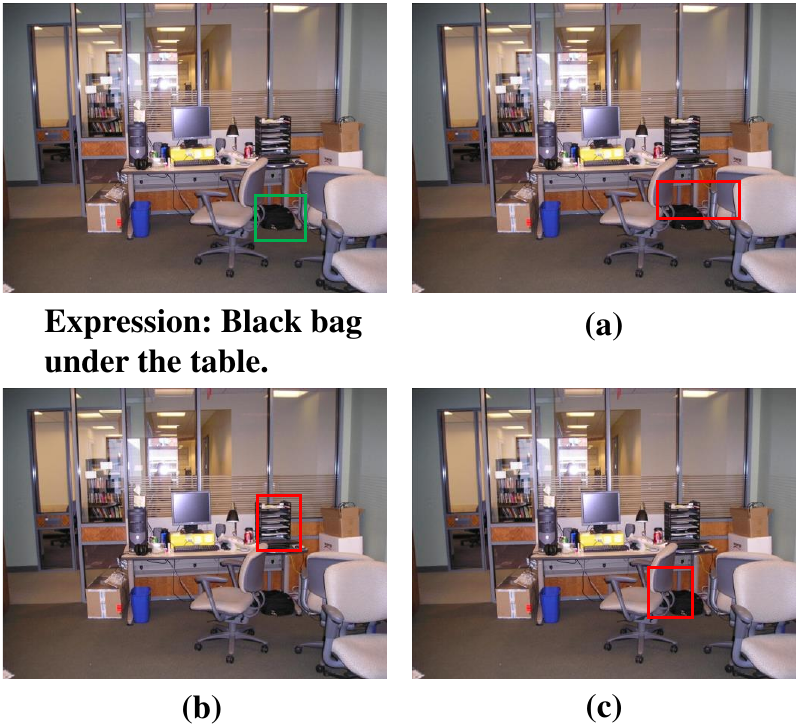}
  %\vspace{-0.10in}
  \caption{Visualizations of the proposed method in different cases. (a), (b) and (c) show the results of our model without EGR, our model without DGC, and our network, respectively.}
  \label{fig:visual} 
\end{figure}

%\vspace{-0.2in}
\subsection{Evaluation Results}
% \paragraph{Evaluation.} 
In the setting of only using the detected objects, we show the evaluation results of the proposed and other compared methods on the RefCOCO, RefCOCO+, RefCOCOg, Flickr30K, Refclef and Ref-reasoning datasets in Table~\ref{tab:total1} where TransVG++~\cite{deng2022transvg++}, OFA-Base~\cite{wang2022ofa}\footnote{There are five different OFA models, and we choose the medium one, OFA-Base, for comparison whose model size is closer to the proposed method, and we both use the same network backbone (\textit{i.e.}, ResNet-101) for visual feature extraction.}, VLTVG~\cite{Yang_2022_CVPR}, QRNet~\cite{ye2022shifting}, and MDETR~\cite{Kamath_2021_ICCV} are the latest transformer-based REC methods which are all pre-trained on a large-scale dataset followed by fine-tuning the models for the REC task. 
%
% The results are reported in Table~\ref{tab:total1} where TransVG++, OFA-base, VLTVG, QRNet, MDETR and OFA-base are the latest transformer-based REC methods where they are pretrained using a large-scale dataset followed by fintuning the models using the REC dataset.
%
% \textcolor[rgb]{1,0,0}{
Table~\ref{tab:total1} shows that our method outperforms the competing transformer-based and non-transformer-based methods in most cases, especially for the testB set of RefCOCO (\textit{i.e.}, comprising more % indecomposable 
short expressions) and the test set of Flickr30K entities where our method respectively reaches much better performance than OFA-Base by $4.55\%$ and MDETR by $3.03\%$. In addition, the proposed method performs $1.05\%$ better than OFA-base on the test set of RefCOCOg (\textit{i.e.}, comprising more complex expressions). Although our method achieves slightly lower performance than OFA-Base on the testA set of RefCOCO and the testB set of RefCOCO+, the proposed model does not require any pre-training on a large-scale dataset and is more compact than OFA-Base. This makes it still a competitive approach with OFA-Base and other transformer-based methods. Moreover, we find the main issue causing inferior performance on these two sets because the employed detector fails to detect proper candidate objects containing the target object specified by the expression (\textit{e.g.}, especially for the testB set of RefCOCO+, most of the target objects are occluded.), and we believe using a better object detector can further improve our performance.

On the other hand, among the compared methods, Ref-NMS~\cite{Chen_Ma_Xiao_Zhang_Chang_2021} and M-DGT~\cite{Chen_2022_CVPR} are the closest approaches to ours. Ref-NMS also introduces a general module to remove the unrelated candidate objects, while M-DGT employs candidate-object pruning as a post-processing refinement by focusing solely on the local graph layout after each message-passing step. However, %as described in Section~\ref{sec:related}, 
since our DGC module is guided by sub-expressions that involve fewer entities and the categorical graph serves as an effective proxy to better align the visual and text domains, it can better match the relevant objects than Ref-NMS and M-DGT. Together with the proposed EGR to refine the predicted boxes, our method significantly outperforms Ref-NMS and M-DGT, as shown in Table~\ref{tab:total1}. Moreover,  Table~\ref{tab:total1} shows the performances of three graph-based methods DGA~\cite{Yangsibei_2019_ICCV}, CMRIN~\cite{8999516}, and SGMN~\cite{Yang_2020_CVPR} are improved evidently when employing the DGC module, further demonstrating the effectiveness of the proposed approach.

Table~\ref{tab:total1} shows a similar trend to the previous results, where the proposed method outperforms the others significantly in most cases. In addition, the proposed DGC module also consistently improves the performances of other graph-based REC methods. These results provide additional evidence of the effectiveness of the proposed approach.

\begin{figure*}[t]
  \centering
  \includegraphics[width = 0.9\textwidth]{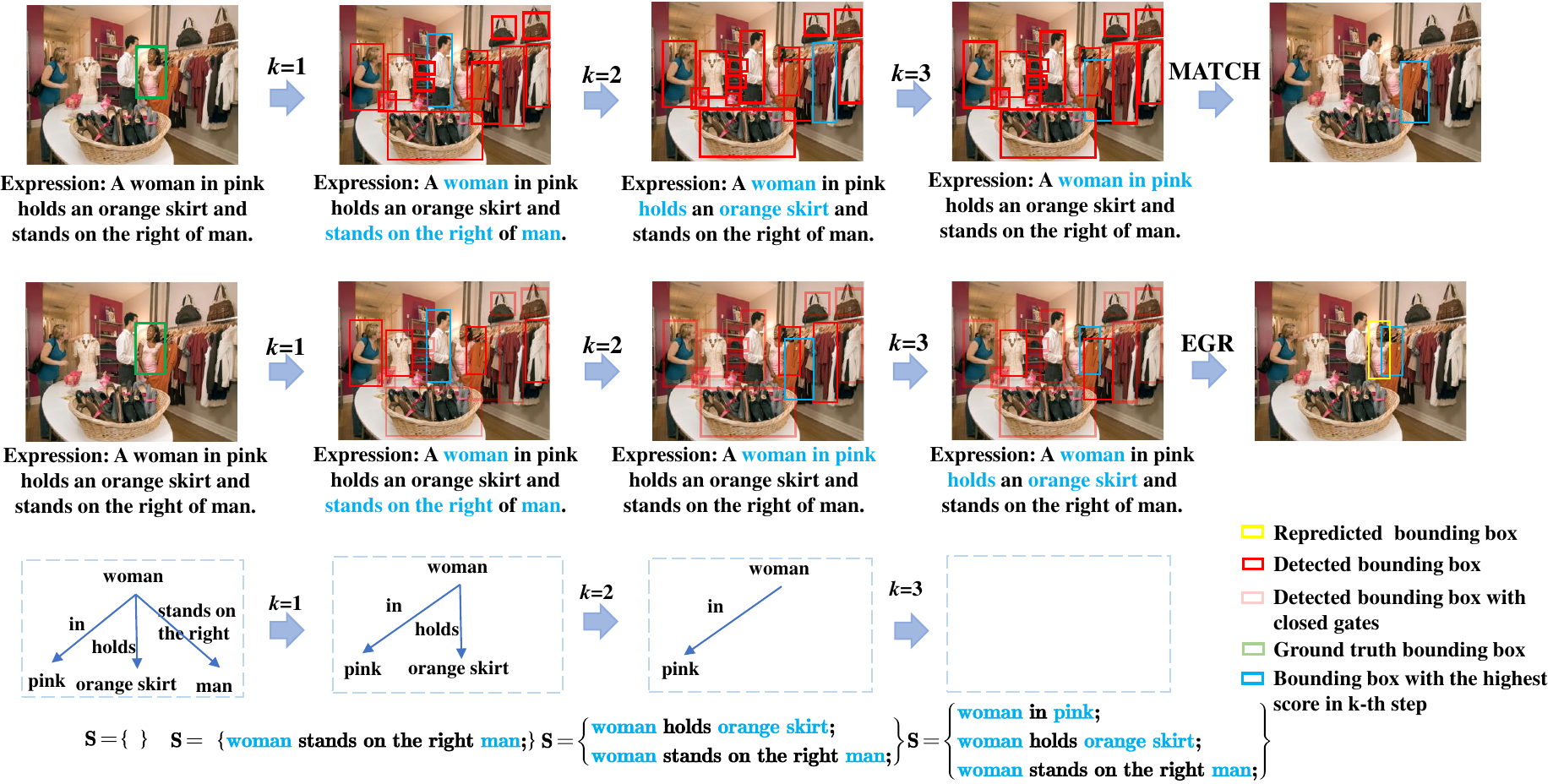}
  %\vspace{-0.10in}
  \caption{Qualitative results of the proposed method. The first, second, and third rows show the results of our method without DGC and EGR, our method, and the processing sequence of sub-expressions, respectively. The words in blue represent the currently processed sub-expression.}
  \label{fig:qualify} 
  %\vspace{-0.12in}
\end{figure*}

\begin{figure}[t]
  \centering
  \includegraphics[width = 0.39\textwidth]{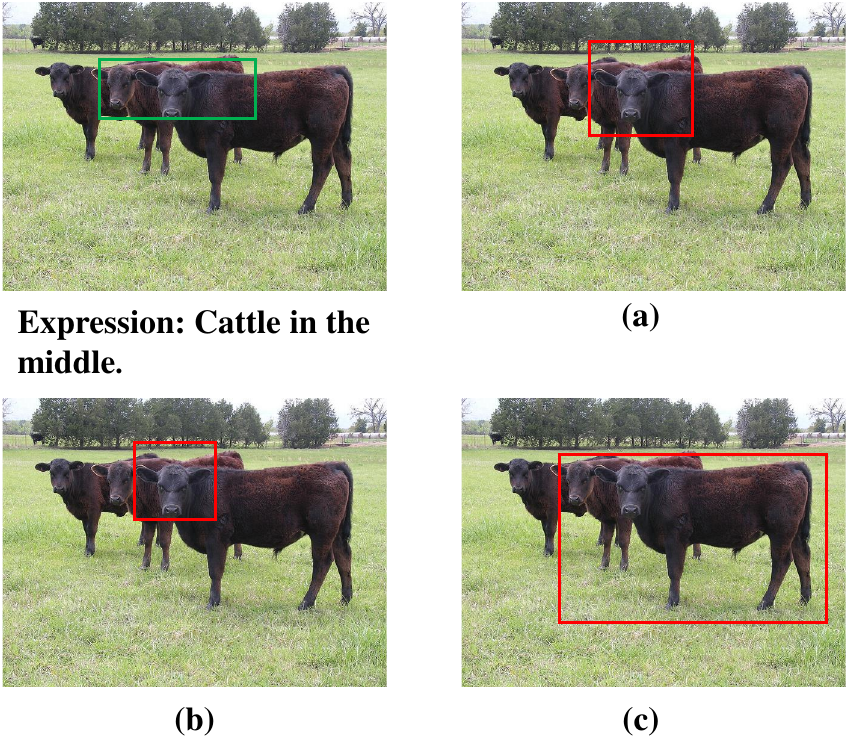}
  %\vspace{-0.10in}
  \caption{Visuals of the proposed method in different cases. (a), (b) and (c) show the results of our model without DGC and our model without both EGR and our network, respectively.}
  \label{fig:error} 
  %\vspace{-0.10in}
\end{figure}

% \begin{table}[t]
% \tiny
% \caption{Ablation studies of our method when only DGC module or EGR strategy is used. D and E represent DGC and EGR, respectively.
% %Ablation results of our method under two strategies of processing sub-expressions.
% }
% %\vspace{-0.05in}
% \label{tab:ablation3}
% \centering
% \renewcommand\arraystretch{1}{
% \setlength{\tabcolsep}{1.5mm}{
% \resizebox{0.47\textwidth}{!}{
% \begin{tabular}{cccccccccc}
% \hline
% % \toprule
% \multirow{2}{*}{D}
% &\multirow{2}{*}{E}
% &\multirow{2}{*}{D+E}
% &\multicolumn{2}{c}{RefCOCO} 
% &\multicolumn{2}{c}{RefCOCO+} 
% &\multicolumn{1}{c}{RefCOCOg}
% &\multicolumn{1}{c}{Ref-reasoning}\\
% \cline{4-9}
% \multicolumn{1}{c}{}
% &\multicolumn{1}{c}{}
% & &testA&testB&testA&testB&test&test\\
% \hline
% \multicolumn{1}{c}{\ding{55}} %\usym{2718} \ding{56}
% &\multicolumn{1}{c}{\ding{55}}
% &\ding{55}&83.96&83.58&74.56&62.54&73.68&51.63\\
% \multicolumn{1}{c}{\checkmark}
% &\multicolumn{1}{c}{}
% & &87.82&85.35&83.58&65.03&78.09&58.76\\
% \multicolumn{1}{c}{} 
% &\multicolumn{1}{c}{\checkmark}
% & &86.67&85.83&76.28&65.35&75.56&55.65\\
% \multicolumn{1}{c}{} 
% &\multicolumn{1}{c}{}
% & \checkmark&\textbf{89.93}&\textbf{87.85}&\textbf{88.16}&\textbf{73.02}&\textbf{83.36}&\textbf{62.64}\\
% \hline
% \end{tabular}}}}
% %\vspace{-5pt}
% \end{table}
\begin{table}[!t]
\caption{Evaluation results of using the proposed DGC in different orders on RefCOCO, RefCOCO+, and RefCOCOg.}
\label{tab:order_later}
\centering
\begin{tabular}{cccccc}
\hline
%\toprule
\multirow{2}{*}{Methods}
% &\multirow{2}{*}{}
&\multicolumn{2}{c}{RefCOCO}
&\multicolumn{2}{c}{RefCOCO+} 
&\multicolumn{1}{c}{RefCOCOg} 
\\
\cline{2-6}
\multicolumn{1}{c}{}
&testA&testB&testA&testB&test\\
\hline
\multicolumn{1}{c}{Forward}
&88.82&\textbf{88.02}&87.84&72.83&82.97\\
% \multicolumn{1}{c}{} 
% &\checkmark&92.78&95.07&90.68\\
\multicolumn{1}{c}{Backward}
&\textbf{89.93}&87.85&\textbf{88.16}&\textbf{73.02}&\textbf{83.36} \\
% \multicolumn{1}{c}{$SSE$}
% &88.37&86.53&87.75&72.36&82.83 \\
% \multicolumn{1}{c}{$ASE$}
% &89.04&87.79&88.10&72.52&83.08 \\
\hline
\end{tabular}
% \vspace{-4pt}
\end{table}

% \begin{table}[t]
% \caption{Comparisons between our method and existing graph-based and transformer-based methods on model size (GB), total parameter size (MB), and inferring time (sec).
% %the model size, parameter size and inferring time of our method and other graph-based models.
% }
% \label{tab:ablation6}
% \centering
% \renewcommand\arraystretch{1}{
% \setlength{\tabcolsep}{3mm}{
% \resizebox{0.45\textwidth}{!}{
% \begin{tabular}{ccccc}
% \hline
% % \toprule
% Method & Model Size & Total Param Size & Inference Time \\
% \hline
% DGA & 0.76 & 145 & 0.38  \\
% CMRIN &\textbf{0.69} &83&0.35 \\
% SGMN&0.73 &\textbf{77}&0.35  \\
% MDETR&3.00&254&1.57  \\
% OFA-base&2.40&180&1.35  \\
% Ours&0.72 &79&\textbf{0.27} \\
% \hline
% \end{tabular}}}}
% \end{table}
\begin{table}[t]
%\tiny
\caption{Comparisons between our method and existing graph-based and transformer-based methods on model size (GB), total parameter size (MB), and inferring time (sec) on RefCOCOg, where the lowest-complexity values are highlighted in bold, and the second-lowest ones are underlined.
}
\label{tab:ablation6}
\centering
\begin{tabular}{cccc}
\hline
% \toprule
Method & Model Size & Total Param Size & Inference Time \\
\hline
DGA & 0.76 & 145 & 0.68  \\
CMRIN &\textbf{0.69} & 83 & 0.62 \\
SGMN&0.73 & \textbf{77} & \underline{0.59}  \\
MDETR & 3.00 & 254 & 1.62  \\
OFA-base & 2.40 & 180 & 1.47  \\
Ours & \underline{0.72} & \underline{79} & \textbf{0.53} \\
\hline
\end{tabular}
\end{table}

% \begin{table*}[t]
% \caption{Comparisons of the model size, parameter size and inferring time of our method and other transformer-based models.}
% %\vspace{-0.05in}
% \label{tab:ablation5}
% \centering
% \renewcommand\arraystretch{1}{
% \setlength{\tabcolsep}{3mm}{
% \resizebox{0.7\textwidth}{!}{
% \begin{tabular}{cccc}
% \hline
% % \toprule
% Method & Model Size(GB) & Total param size(MB) & Inference times(sec) \\
% \hline
% MDETR&3.00&254&1.57  \\
% OFA-base&2.40&180&1.35  \\
% Ours&\textbf{0.72}&\textbf{79}&\textbf{0.27} \\
% \hline
% \end{tabular}}}}
% %\vspace{-15pt}
% \end{table*}
%%
%\vspace{-0.04in}
\subsection{Ablation Studies}
We conduct various ablation studies to verify the effectiveness of different components of the proposed method. We evaluate our REC models on the RefCOCO, RefCOCO+, and RefCOCOg datasets
%, where RefCOCO and RefCOCO+ have expressions with an average length of $3.5$ words, while RefCOCOg has longer expressions with an average of $8.4$ words. Moreover, the proportion of short expressions (\textit{i.e.}, containing only one or two noun chunks) in RefCOCO, RefCOCO+, and RefCOCOg is $85.26\%$, $87.73\%$, and $46.77\%$, respectively.

We first study the performance influences of different detectors on the proposed method. We employ Faster R-CNN and train different object detectors on the VG, COCO, and VOC datasets, respectively. We evaluate our REC models on  RefCOCO, RefCOCO+, and RefCOCOg. Table~\ref{tab:ablationde} shows that although better detection results could help improve the performance, the proposed method is robust and achieves satisfactory results even when the object detector does not perform well. Moreover, since VG contains many annotation errors and noise, it thus results in inferior performance when involving it in training (see the third row in Table~\ref{tab:ablationde}).

Based on the Faster R-CNN trained on the COCO dataset,  we validate the effectiveness of our method when using the proposed DGC and EGR modules, respectively. The results are reported in Table~\ref{tab:ablation3} and Fig.~\ref{fig:visual}. From Table~\ref{tab:ablation3}, we find that using either DGC or EGR can boost the performance of the baseline model, and using both simultaneously leads to the best performance. Meanwhile, combining the results from Table~\ref{tab:ablation3} and Table~\ref{tab:total1}, it is evident that the baseline (our method without DGC and EGR) outperforms the majority of graph-based methods. Moreover, our method with only DGC surpasses DGA with DGC, CMRIN with DGC, and SGMN with DGC. The results show that our novel node and edge weight generation strategy in (4)--(8) enables a more precise establishment of relationships between entities in sub-expressions and nodes in the two graphs during reasoning than using the entire expression employed by other methods. As shown in Fig.~\ref{fig:visual}, we can find that when only using either DGC or EGR, the proposed method fails to locate the target. However, when using both, the proposed method can successfully locate the target object.

Since the REC benchmark datasets do not offer the structural labels of the expressions, there is no ground truth to evaluate the adopted language parser UniVSE~\cite{Wu_2019_CVPR} on these benchmarks.
To address this issue, we manually label a set of expressions and evaluate UniVSE.
Specifically, we conduct an experiment using the RefCOCOg and Ref-reasoning datasets, which are known for containing longer expressions. 
The numbers of expressions in RefCOCOg and Ref-reasoning are 95,010 and 142,956, respectively. 
It is almost infeasible for us to label all of them. 
Therefore, we randomly select 500 expressions from each dataset and follow the approach used for labeling sentences in~\cite{Wu_2019_CVPR} to annotate the word dependencies for each expression manually.
These selected expressions and the corresponding annotations serve as the ground truth to evaluate UniVSE. 
The results reported in Table~\ref{tab:parser} illustrate that UniVSE can accurately parse most of the expressions selected from the RefCOCOg and Ref-reasoning datasets. 
By comparing the performance shown in the first two rows of Table~\ref{tab:ablation3} of the paper, it can be observed that the proposed DGC module, working on the parsed expressions by UniVSE, substantially improves performance on all the datasets.
The results reveal that the UniVSE parser indeed helps, even though it is not perfect.
% %
% We randomly selected 500 expressions from each dataset and followed the approach used for labeling sentences in~\cite{Wu_2019_CVPR} to annotate the word dependencies for each expression manually. 
% %
% These selected expressions and the corresponding annotations serve as the ground truth to evaluate UniVSE. 
% %
% The results reported in Table~\ref{tab:parser} illustrate that UniVSE can accurately parse most of the expressions selected from the RefCOCOg and Ref-reasoning datasets. 
% %
% By comparing the performance shown in the first two rows of Table~\ref{tab:ablation3}, it can be observed that the proposed DGC module, working on the parsed expressions by UniVSE, substantially improves performance on all the datasets, even though the UniVSE parser is not perfect.

Next, we use ${\cal G}_{a}$, ${\cal G}_{c}$ and ${\cal G}_{a}$+${\cal G}_{c}$ to verify the performance of our method.
Table~\ref{tab:ablation2} shows that the performance of using both ${\cal G}_{a}$ and ${\cal G}_{c}$ is better than that using either of them, revealing that the two graphs are complementary to each other. 
Meanwhile, the performance of using only ${\cal G}_{a}$ is better than that of ${\cal G}_{c}$ since ${\cal G}_{a}$ contains richer information than ${\cal G}_{c}$ (\textit{i.e.}, the dimension of node features of ${\cal G}_{a}$ is $2,048$ while ${\cal G}_{c}$ is $300$). 
This confirms the necessity and effectiveness of using both graphs to provide more context information for the REC task.  

Together with the results of DGA with DGC, SGMN with DGC, and CMRIN with DGC in Table~\ref{tab:total1}, it demonstrates that DGC is a general module and suitable for different graph-based REC methods and the EGR strategy helps overcome the problem of inaccurate detection by detector.
Moreover, from both Table~\ref{tab:ablation3} and Table~\ref{tab:ablation2}, we can also find that although for RefCOCO and RefCOCO+, there are many short and indecomposable expressions, our bimodal graph interaction in the DGC module along with EGR can still help extract the essential information from them and benefit the reasoning performance effectively.

Furthermore, we investigate the impact of the processing ordering of each sub-expression. As shown in Table~\ref{tab:order_later}, the performances in the forward and backward orders are close, which shows the processing order does not cause a significant impact on performance.
This demonstrates that the proposed method is not sensitive to any processing order of using sub-expressions to control the gates. For the best performance in most cases, we use the results of using the backward processing order for the whole paper.

Finally, we validate the complexity of our method,  which includes both the Faster R-CNN detector and the proposed graph-based inference modules, by comparing the total model size, the number of parameters, and the inference time with other three SOTA graph-based REC methods and two transformer-based methods. The machine for evaluation is equipped with one NVIDIA 2080Ti GPU and one Intel i7-9800X CPU. We evaluate our method using all test samples from  RefCOCOg, which contains longer expressions. The average inference times are presented in Table~\ref{tab:ablation6}. We can see that our method with DGC has a comparable model size with the other SOTA graph-based methods. Moreover, our approach achieves faster inference speed than the competing graph-based methods since the proposed DGC allows us to achieve the desired performance in fewer reasoning steps. In comparison to transformer-based methods, it shows that the model and parameter sizes of the proposed method are much smaller than MDETR and OFA-base by at least $1.68$~GB and $101$~MB, respectively. In contrast, the inference time of our method is faster than MDETR by $1.09$ seconds and OFA-Base by $0.94$ seconds.
\\

%\vspace{-0.1in}
\subsection{Qualitative Results}
%\vspace{-0.08in}
Besides the quantitative results, we also show the visuals of the proposed method in 
Fig.~\ref{fig:qualify}.
% displays some qualitative results of the proposed method. 
%
The first, second, and third rows illustrate the results of our method without DGC and EGR, our method, and the processing sequence of sub-expressions during the multi-step reasoning phase, respectively.
From Fig.~\ref{fig:qualify}, with the help of the DGC module, the proposed method can effectively reduce the influences from the irrelevant objects to the expression. Furthermore, our method can refine the bounding box with the EGR module to better locate the target object, as shown in the second row. These results demonstrate the effectiveness of the proposed framework with the DGC and EGR modules.
Nevertheless, Fig.~\ref{fig:error} shows a failure case where our method can correctly locate the cattle while failing to predict the correct bounding box due to the target object being surrounded by many similar objects. Together with the results of DGA with DGC, SGMN with DGC, and CMRIN with DGC in Table~\ref{tab:total1}, they demonstrate that the proposed DGC module can effectively disable the nodes unrelated to the expression and further helps the EGR module predict a more precise bounding box.

\section{Conclusion}
In this paper, we showed the proposed DGC is a general plug-and-adapt module that can effectively boost the performances of various graph-based REC models by mitigating the interference from those object candidates and connections irrelevant to a given expression during the reasoning stage. 
Furthermore, we have shown our EGR module can further help the proposed framework locate the target object described by the expression more precisely.
Besides applying DGC and EGR, the proposed graph-based REC method with both visual and categorical graphs can further outperform the current state-of-the-art transformer-based REC methods on most REC datasets without the need for pre-training on large-scale datasets and exploiting complex architectures. 
With these advantages, our method makes the graph-based REC method a powerful tool again for the REC task.

\ifCLASSOPTIONcaptionsoff
  \newpage
\fi

% trigger a \newpage just before the given reference
% number - used to balance the columns on the last page
% adjust value as needed - may need to be readjusted if
% the document is modified later
%\IEEEtriggeratref{8}
% The "triggered" command can be changed if desired:
%\IEEEtriggercmd{\enlargethispage{-5in}}

% ====== REFERENCE SECTION

%\begin{thebibliography}{1}

% IEEEabrv,

\bibliographystyle{IEEEtran}
\bibliography{IEEEabrv,rec}

\vfill

% Can be used to pull up biographies so that the bottom of the last one
% is flush with the other column.
%\enlargethispage{-5in}

\clearpage

% that's all folks
\end{document}